\documentclass[10pt,twocolumn,letterpaper]{article}

\usepackage{cvpr}              

\usepackage{microtype}
\usepackage{nicefrac}
\usepackage[american]{babel}
\definecolor{cvprblue}{rgb}{0.21,0.49,0.74}
\usepackage[pagebackref,breaklinks,colorlinks,allcolors=cvprblue]{hyperref}

\def\paperID{63} 
\def\confName{CVPR}
\def\confYear{2026}

\title{ArtiWorld: LLM-Driven Articulation of 3D Objects in Scenes}

\author{
    Yixuan Yang$^{1}$\footnotemark[1]\quad
    Luyang Xie$^{1}$\footnotemark[1]\quad
    Zhen Luo$^{1,2}$\footnotemark[1]\quad
    Zixiang Zhao$^{3}$\\
    Tongsheng Ding$^{1}$\quad
    Mingqi Gao$^{1}$\quad
    Feng Zheng$^{1,4}$\footnotemark[2]
    \\[1mm]
    $^{1}$SUSTech \quad
    $^{2}$SII\quad
    $^{3}$ETH Z\"urich\quad
    $^{4}$Spatialtemporal AI \\
     \tt \small arnoldyang97@gmail.com, 12532566@mail.sustech.edu.cn, luoz2024@mail.sustech.edu.cn \\
}

\begin{document}
\maketitle
\begin{abstract}
Building interactive simulators and scalable robot-learning environments requires a large number of articulated assets. However, most existing 3D assets in simulation are rigid, and manually converting them into articulated objects is extremely labor- and cost-intensive. This raises a natural question: can we automatically identify articulable objects in a scene and convert them into articulated assets directly?
In this paper, we present \textbf{ArtiWorld}, a scene-aware pipeline that localizes candidate articulable objects from textual scene descriptions and reconstructs executable URDF models that preserve the original geometry.
At the core of this pipeline is \textbf{Arti4URDF}, which leverages 3D point cloud, prior knowledge of a large language model (LLM), and a URDF-oriented prompt design to rapidly convert rigid objects into interactive URDF-based articulated objects while maintaining their 3D shape. 
We evaluate ArtiWorld at three levels: 3D simulated objects, full 3D simulated scenes, and real-world scan scenes. Across all three settings, our method consistently outperforms existing approaches and achieves state-of-the-art performance, while preserving object geometry and correctly capturing object interactivity to produce usable URDF-based articulated models. This provides a practical path toward building interactive, robot-ready simulation environments directly from existing 3D assets.
Code and data will be released.

\end{abstract}    
\vspace{-1em}
\section{Introduction}
\label{sec:intro}
\begin{figure*}[t]
\vspace{-1em}
    \centering
    \includegraphics[width=0.9\linewidth]{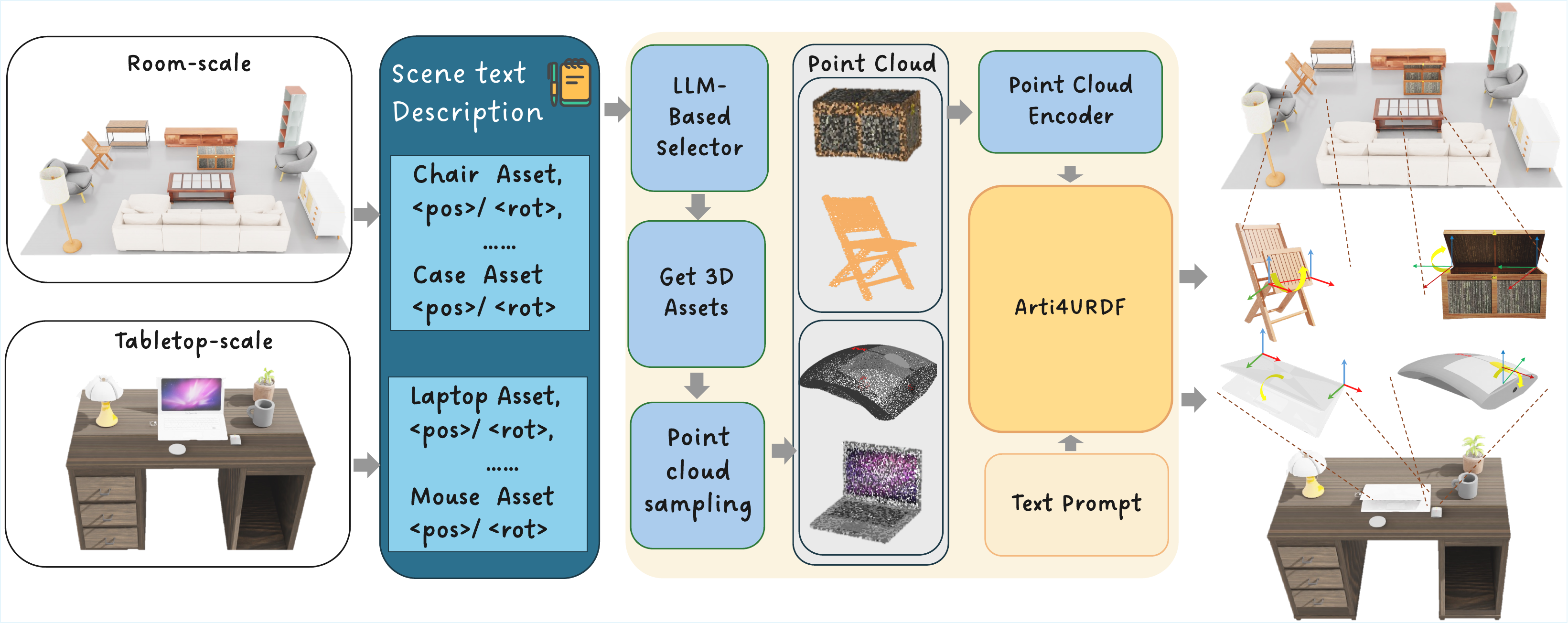}
        \vspace{-1em}
    \caption{ \textbf{Overview of ArtiWorld pipeline.}
Given a room-scale or tabletop-scale scene description, our system uses a language model to identify objects that should be articulated and retrieves their corresponding 3D assets. Point cloud surfaces are sampled from each object and encoded as geometric tokens. These tokens, together with a structured text prompt, are fed into Arti4URDF, which predicts joint types, axes, and articulation limits. The generated URDF models are then aligned back to their original scene positions, producing fully interactive articulated scenes suitable for simulation and downstream robotic tasks.}
    \label{fig:arti_teaser}
        \vspace{-1.2em}
\end{figure*}

Converting rigid 3D assets into articulated objects in simulation is essential for scaling data and improving robot learning~\cite{ren2024infiniteworld,makoviychuk2021isaac,dimitropoulos2022brief,hanna2017grounded}. 
However, collecting large-scale interactive articulated 3D data in the real world is expensive and requires substantial manual effort, which makes simulation-based data augmentation a more practical alternative. Large repositories of static 3D objects, such as Objaverse~\cite{deitke2023objaverse}, 3D-Front~\cite{fu20213d}, InternScenes\cite{zhong2025internscenes} and ShapeNet~\cite{chang2015shapenet}, are already available. Yet many scene generation methods built on these datasets, including MesaTask~\cite{hao2025mesatask} and Holodeck~\cite{yang2024holodeck}, still produce scenes where most objects remain static and cannot support meaningful robot interaction. This leads to a natural question: can we directly extract rigid objects from existing simulated scenes and automatically convert them into articulated, geometry-preserving assets that robots can immediately interact with?

We begin our discussion from the perspective of simulated scenes. Representative scene generation works~\cite{yang2024llplace,yang2024physcene,yang2025sceneweaver,yang2025optiscene,hao2025mesatask} such as Holodeck~\cite{yang2024holodeck}, OptiScene~\cite{yang2025optiscene}, and PhyScene~\cite{yang2024holodeck} synthesize and arrange furniture at the room scale, while methods like MesaTask~\cite{hao2025mesatask} focus on tabletop-scale simulation data. However, these pipelines typically treat all objects as rigid and do not explicitly identify or parameterize objects that should be articulable, such as doors, drawers, or cabinet fronts. Although PhyScene~\cite{yang2024physcene} and MesaTask~\cite{hao2025mesatask} can insert articulated objects by retrieving them from external datasets, they still depend on pre-existing articulated assets rather than converting the rigid objects that are already present in the scene. This leaves a persistent gap between scene generation and interactive robot learning. Our goal is to build a unified pipeline that detects and localizes articulable objects at the scene level and then reconstructs their articulations, enabling existing simulated scenes to become truly interactive.

In parallel, at the single-object level, articulated modeling has also attracted increasing attention. Methods such as CAGE~\cite{liu2024cage} and Infinity Mobility~\cite{lian2025infinite} construct articulated objects by retrieving parts and defining their connections with graph-based structures, which restricts category diversity and often leads to homogenized designs. Articulate-Anything~\cite{articulate-anything} still relies on part retrieval but uses large language models to predict linkages and joint types between components; although more expressive, the retrieved parts are typically simple and the assembled objects often suffer from collisions or misalignment. Another line of work, including Articulate-AnyMesh~\cite{qiu2025articulate}, starts from a 3D mesh object and uses vision-language models to infer joint positions and axes. URDFormer~\cite{chen2024urdformer} trains a transformer-based model that converts objects into simplified geometric primitives, which makes it difficult to preserve the original object shape.
These method pipelines struggle to maintain the full 3D geometry of objects, which can result in inaccurate or ambiguous contact points and articulation parameters.
Recent work URDF-Anything~\cite{urdfanything} takes the entire object point cloud as input and predicts both semantic part segmentation and URDF files using a multimodal LLM as prior knowledge. Although this approach preserves the original geometry, the multi-task learning setup inevitably introduces errors in segmentation and articulation prediction.

To tackle the above challenges, in this paper, as shown in \cref{fig:arti_teaser}, we introduce \textbf{ArtiWorld}: given a scene description (textual and/or geometric), we first localize candidate articulable objects and extract their 3D structure. 
We then introduce \textbf{Arti4URDF} model, a generative model that predicts joint topology and kinematic parameters and outputs executable URDFs (Unified Robot Description Format). Our goal is to convert existing non-articulated 3D assets into interactive articulated objects in URDF while preserving their original geometry.
Arti4URDF integrates 3D point cloud tokens into a large language model to generate URDF descriptions directly. 
It takes a sequence of encoded point cloud tokens as input and produces both high-level structural relationships and low-level URDF syntax.
During training, we leverage a combined dataset with PartNet-Mobility~\cite{xiang2020sapien} and PhysXNet~\cite{cao2025physx}, which can be assembled as a whole 3D object with some 3D parts.
Each part's point cloud is encoded using the ULIP point cloud encoder to obtain a sequence of point cloud tokens. 
We then construct a 3D-point-cloud-to-URDF prompt that embeds these tokens into the LLM's input. 
The model is fine-tuned to first output a link-joint graph describing inter-part relationships, followed by a detailed URDF file.
At inference time, we observe that existing 3D part segmentation models, such as PartSLIP++~\cite{zhou2023partslip++}, often fail to produce reliable decompositions. So we design a point prompt-based segmentation strategy that selects representative points to guide part decomposition, and the segmented parts are then fed into Arti4URDF for URDF prediction. The predicted URDF structure can be projected back onto the original mesh parts or onto the segmented 3D point cloud parts to obtain articulated 3D assets. Finally, these articulated assets are aligned and re-inserted into their original scene locations, yielding physically interactive, articulation-aware simulated scenes.

Our main contributions are as follows:
(1)~We propose \textbf{ArtiWorld}, a scene-aware pipeline that identifies articulable objects from textual and geometric scene descriptions, converts rigid assets in existing simulated scenes into articulated URDF-based objects that can be directly used for robot interaction.
(2)~We introduce \textbf{Arti4URDF}, a novel model that embeds 3D assets shape features into a large language model (LLM) to directly infer inter-part relationships, joint types, and joint positions, and generate complete URDF files in an end-to-end manner.
(3)~We evaluate our method on 3D object-level, 3D scene-level, and real-world scans object test dataset, and demonstrate state-of-the-art (SOTA) performance in both joint type prediction and joint axis localization compared to prior approaches.

\section{Related Work}
\subsection{LLM for 3D Structure Understanding}
Large Language Models (LLMs) such as GPT~\cite{achiam2023gpt}, PaLM~\cite{anil2023palm}, LLaMA~\cite{touvron2023llama}, and Qwen~\cite{bai2023qwen} have demonstrated remarkable capabilities in language understanding and general-purpose reasoning. Recently, research has extended LLMs to multimodal domains, enabling them to process images, audio, video, and even 3D data. In the vision-language space, models like BLIP-2~\cite{li2023blip} and LLaVA~\cite{liu2023visual} integrate image features with language inputs to enable visual question answering, captioning, and reasoning.

However, applying LLMs to 3D data, especially point clouds, remains a nascent area due to the irregularity and sparsity of 3D structures. Recent works such as PointLLM~\cite{xu2024pointllm} and MiniGPT-3D~\cite{tang2024minigpt} explore aligning point cloud encoders with LLMs to perform natural language descriptions, classification, and interaction over 3D objects. PointLLM builds a two-stage framework with a projection layer and instruction tuning, while MiniGPT-3D introduces a lightweight strategy using 2D vision-language models as bridges for 3D-to-text alignment. These methods demonstrate the potential of LLMs for 3D perception, though their focus remains on recognition rather than articulation.

\subsection{Articulated Object Modeling}
Modeling articulated objects—composed of rigid parts connected by joints—is a long-standing challenge in 3D vision and robotics. Traditional approaches~\cite{sharf2014mobility, yuan2016space,mitra2010illustrating,yi2018deep, wang2019shape2motion,kawana2023detection, deng2024articulate} often rely on multi-view images or motion-capture data, which limits scalability.
Recent efforts have attempted to automate articulation modeling.
CAGE~\cite{liu2024cage} and Infinity Mobility~\cite{lian2025infinite} construct articulated objects by retrieving parts and defining their connections with graph-based structures. Ditto~\cite{jiang2022ditto} reconstructs digital twins of articulated objects by analyzing pre- and post-interaction point clouds. SINGAPO~\cite{liu2024singapo} predicts part structure and joint attributes from a single RGB image using part graphs and diffusion-based generation. Articulate-AnyMesh~\cite{qiu2025articulate} converts static meshes into articulated assets by combining semantic segmentation with joint prediction via vision-language models. Articulate-Anything~\cite{articulate-anything} further explores multimodal articulation (text, image, video) using LLMs for joint inference and asset composition in open-vocabulary settings. URDF-Anything~\cite{urdfanything} processes the full object point cloud and uses a multimodal LLM to infer both semantic part decomposition and URDF parameters in a single framework.
While these methods have advanced the field, most rely heavily on 2D modalities and part retrieval. They often suffer from geometry mismatch, ambiguous contact predictions, or limited structural diversity—issues our method explicitly addresses by grounding articulation inference directly in 3D geometry through point cloud token embeddings and URDF generation.

\section{Methodology}
\begin{figure*}
    \centering
    \includegraphics[width=0.95\linewidth]{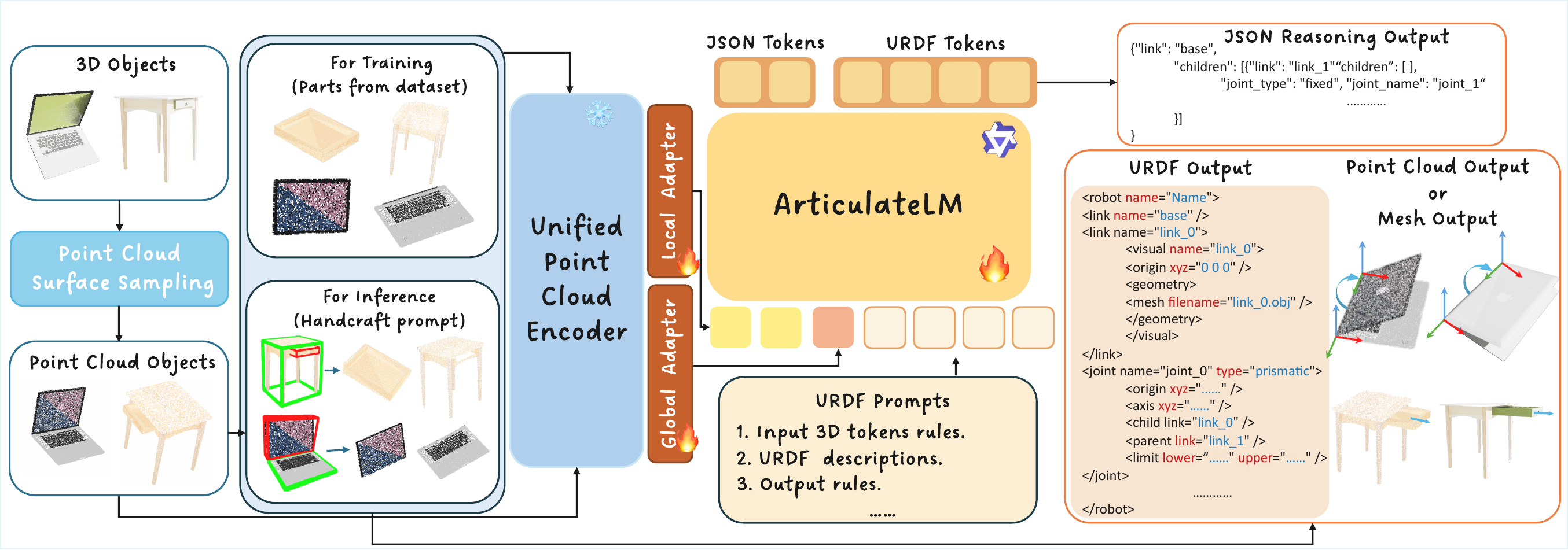}
    \caption{
\textbf{ Overview of the Arti4URDF pipeline.}
Our Arti4URDF takes raw 3D objects and samples their surfaces to obtain point clouds for training and inference. A unified point cloud encoder extracts both global and local geometric features, which are mapped into the LLM embedding space through lightweight adapters. These features are injected into structured URDF prompts that describe part–joint relationships and articulation rules. The LLM-based Arti4URDF model then generates JSON-style structural descriptions and full URDF files, which can be used to produce articulated mesh or point cloud outputs that are ready for simulation and downstream robotic interaction.
}
    \label{fig:main_pipeline}
    \vspace{-1em}
\end{figure*}

In this section, we first introduce the AriWorld (\cref{fig:arti_teaser}) in ~\cref{sec:ArtiWorld}. And then we discuss the Arti4URDF model (\cref{fig:main_pipeline}) to convert 3D assets to URDF in~\cref{sec:Arti4URDF}. Finally, we discuss the training and inference procedures in~\cref{sec:t_i}.

\subsection{ArtiWorld}
\label{sec:ArtiWorld}
First, we define: For the JSON file of a scene description $Scene_t$, we use a LLM to determine which objects can be articulated.
We then output the descriptions of the corresponding articulated objects and their respective asset IDs, so as to obtain the corresponding 3D asset sequence \( \mathcal{O} \in \{O_1, \dots, O_n\} \). After that, for each articulated object \( \mathcal{O} \), we consider that it can be regarded as: Let a 3D object \( \mathcal{O} \) be composed of a set of \( n \) rigid parts:
\begin{equation}\label{}
\small
    \mathcal{O} = \{ p_1, p_2, \ldots, p_n \},
\end{equation}
where each $p_i$ represents a static 3D part (e.g., a point cloud or mesh). Our goal is to transform the static object $\mathcal{O}$ into an articulated version $\mathcal{O}_{\text{art}}$, where the parts are connected by joints with physically meaningful kinematic relationships.

To achieve this, we design Arti4URDF $\mathcal{M}$ that takes the object $\mathcal{O}$ as input and generates a corresponding URDF (Unified Robot Description Format) file $\mathcal{U}$. Arti4URDF consists of a 3D encoder that extracts geometric features from the input point cloud and an LLM-based module, ArticulateLM, which performs structural inference and URDF generation conditioned on these features:
\begin{equation}\label{}
\small
\mathcal{M}(\mathcal{O}) = \mathrm{ArticulateLM}\left(\mathrm{Enc}_{3D}(\mathcal{O})\right) \rightarrow \mathcal{U}.
\end{equation}
The URDF file $\mathcal{U}$ defines the structure of the articulated object, including the set of links (corresponding to parts $p_i$), the joint types (e.g., revolute, prismatic), joint parameters (axes, limits), and the parent-child hierarchy.

Using $\mathcal{U}$, we can then instantiate the articulated object $\mathcal{O}_{\text{art}}$ as follows:
\begin{equation}\label{}
\small
    \mathcal{O}_{\text{art}} = \texttt{Assemble}(\mathcal{O}, \mathcal{U}),
\end{equation}
where $\texttt{Assemble}$ denotes the simulation-level process that binds parts $\{p_i\}$ together according to the joint configurations specified in $\mathcal{U}$. The resulting object $\mathcal{O}_{\text{art}}$ supports interaction, simulation, and control in robotic environments.
After articulating objects, these objects will be placed in the original rigid objects in the scenes (As shown in~\cref{fig:arti_teaser}).

\subsection{Arti4URDF}
\label{sec:Arti4URDF}
\subsubsection{Encode 3D point cloud to ArticulateLM}
In Section \ref{sec:ArtiWorld}, we defined a 3D articulated object as a composition of multiple connected parts. Therefore, a critical prerequisite is obtaining meaningful part-level decomposition from a full 3D asset.
Existing datasets often provide overly fragmented part annotations that include non-functional subdivisions unsuitable for articulation modeling. To address this, we merge over-segmented parts and retain only those that participate in articulation, simplifying the structure while preserving functional joints.
After segmentation, the original object O can be represented as a set of parts
$
\mathcal{O} = \{ p_1, p_2, \ldots, p_n \}.
$

To enable the ArticulateLM to reason over inter-part relationships, it is critical to inject geometric structure into the input. This enhances the model’s capability to understand spatial structure and predict plausible joint types based on part positioning.
Because point-cloud encoders such as \texttt{Point\mbox{-}BERT}~\cite{yu2022point} are pre-trained on \textit{complete, normalized} shapes, directly feeding raw part clouds can introduce scale/translation bias. We therefore normalize each part prior to encoding and then construct a token sequence in a fixed order (e.g., by part index):
\begin{equation}\label{}
\small
\tilde{p}_i=\phi_{\text{norm}}(p_i),
\mathbf{f_{p,i}}=\texttt{Point\mbox{-}BERT}(\tilde{p}_i)\in\mathbb{R}^{d_{\text{geo}}},
\end{equation}
where \(p_i\in\mathcal{O}\) is the \(i\)-th part point cloud, \(\phi_{\text{norm}}(\cdot)\) applies per-part centering and scale normalization, \(\mathbf{g}_i\) is the high-dimensional geometric feature, and \(d_{\text{geo}}\) is the encoder output dimension. The sequence $\mathcal{F}_p=[\,\mathbf{f}_{p,1},\ldots,\mathbf{f}_{p,n}\,]$ serves as the geometric token stream to the LLM, equipping it with shape and local spatial cues essential for predicting link–joint graphs and kinematic parameters.

For our task, beyond enabling the model to understand each part’s geometry as a local geometric prior—which aids in predicting the precise rotation axis and joint type—we also want the model to capture the global structure of the entire object. Therefore, we additionally encode the complete object point cloud as a global geometric prior to help the model reason about the spatial relationships among all parts.
Beyond local priors, we also want the model to capture the \textit{global} structure of the entire object to reason about inter–part spatial relations. Hence, we encode the complete object point cloud as a global geometric prior $\mathbf{f}_g$:
\begin{equation}\label{}
\small
\mathbf{f}_g=\texttt{Point\mbox{-}BERT}(\mathcal{O})\in\mathbb{R}^{d_{\text{geo}}}.
\end{equation}
Final 3D input token sequence is
$
\mathcal{F} = \bigl[\mathbf{f}_g,\ [\mathbf{f}_{p,1},\ldots,\mathbf{f}_{p,n}]\,\bigr]
$,
where \(\mathbf{f}_g\) provides the global geometric context and \(\{\mathbf{f}_{p,i}\}\) supply local, part–level conetext.

To align the feature dimension \(d_{\text{geo}}\) of the 3D point-cloud encoder with the token embedding dimension \(d_{\text{ArticulateLM}}\) of the ArticulateLM, we introduce a lightweight two-layer MLP as a 3D-to-LLM adapter. Given a geometric feature \(\mathbf{f}\in\mathbb{R}^{d_{\text{geo}}}\), the adapter produces
$
\mathbf{t}=\texttt{MLP}(\mathbf{f})\in\mathbb{R}^{d_{\text{ArticulateLM}}},
$
where \(\mathbf{t}\) denotes the transformed token corresponding to a 3D point-cloud feature. The adapter projects geometric features into the same latent space as the LLM tokens while preserving the semantic and geometric structure of the parts. For the global feature and the part features, we apply separate adapters to obtain the global token \(\mathbf{t}_g\) and the part-token set \(T=\{\mathbf{t}_1,\mathbf{t}_2,\ldots,\mathbf{t}_n\}\), which are then injected into the LLM’s prompt template. By conditioning the ArticulateLM on \(\{\mathbf{t}_g\}\cup T\), the model can reason about spatial relationships, joint types, and kinematic hierarchies based on the encoded 3D geometry.

\noindent \textbf{Prompt for point cloud token inputs.}
We first describe how we represent the point-cloud input tokens within the prompt.
We craft a hierarchical prompt that injects the full-object and per-part point cloud embeddings into the ArticulateLM context using special boundary tokens. The full-object embedding appears first between $<g\_start>$ and $<g\_end>$ to convey global scale and layout. The per-part embeddings are concatenated in the exact order of the provided filenames between $<p\_start>$ and $<p\_end>$ (i.e., $link_0\rightarrow… \rightarrow link_N$), ensuring unambiguous correspondence between embeddings and link names. Listing the object name and the part filenames in the prompt further eliminates permutation ambiguity, enabling the model to leverage both global context (for base-link identification and relative placement) and local cues (contacts, symmetries, elongation) for structural reasoning.

\subsubsection{Structural Reasoning Chain}
Transforming 3D geometry into a structured URDF representation is challenging because it requires reasoning over part relationships as well as generating precise motion parameters and constraints. This cross-domain mapping—from unstructured geometry to symbolic kinematic specification—benefits from an intermediate representation that bridges geometric perception and articulation modeling.

To this end, we introduce a structural reasoning chain that decomposes URDF generation into explicit, interpretable steps. The chain begins by forming a coarse structural hypothesis that groups parts and infers their parent–child connections, yielding a preliminary kinematic tree based on geometric cues such as adjacency and spatial contact.

This structured reasoning process acts as a prior that links part-level point cloud information to the final URDF. By explicitly encoding intermediate relational hypotheses (e.g., contact, linkage, potential motion axes), the model can progressively refine geometric evidence into physically consistent and simulator-ready articulation specifications.

Formally, the reasoning chain is serialized as an ordered sequence of inference steps that culminate in two delimited outputs: a JSON block and a URDF block.

\begin{itemize}
\item The first is a JSON block, enclosed between \texttt{<json\_start>} and \texttt{<json\_end>}, specifying the kinematic tree with the fields \textit{part}, \textit{joint\_type}, and \textit{children} (optional \textit{joint\_name}), while omitting all numeric parameters.
\item The second is a URDF block, enclosed between \texttt{<urdf\_start>} and \texttt{<urdf\_end>}, containing a complete URDF description: each \textit{link} is named using its mesh filename and includes a \textit{visual/mesh} reference; each \textit{joint} defines its \textit{type/parent/child} and, for movable joints, the \textit{axis/limit} with a default zero origin. Virtual helper links are added for compound motions. \textit{Inertial} and \textit{collision} terms are omitted, and all \textit{link} elements are declared before their use in \textit{joint} elements.
\end{itemize}

The prompt instructs the model to act as a domain expert in URDF modeling and mechanical structure analysis. It provides:
\begin{itemize}
\item \textbf{Reasoning Instructions:} The model infers the parent–child hierarchy, identifies the base link, and determines joint types (\texttt{revolute}, \texttt{prismatic}, \texttt{fixed}) from geometry, contact, symmetry, and possible motion, inserting helper links when compound joints are present.
\item \textbf{Output Constraints:} The model must output (i) a JSON kinematic tree delimited by \texttt{<json\_start>} and \texttt{<json\_end>}, and (ii) a complete URDF file delimited by \texttt{<urdf\_start>} and \texttt{<urdf\_end>}, both strictly following URDF syntax for simulator compatibility.
\end{itemize}

This structured prompt provides both geometric context and task-specific guidance, enabling the LLM to generate physically consistent, simulator-ready articulated object descriptions.

\subsection{Training and Inference}
\label{sec:t_i}
Our model is trained to generate both the Structural Reasoning Chain and the corresponding URDF parameters directly from 3D point cloud inputs using a language modeling loss. During training, the model consumes point clouds from predefined datasets, where each object is already segmented and annotated for articulation, enabling learning of consistent mappings between part-level geometry and kinematic structure. The model autoregressively predicts the entire structured output sequence, with both the reasoning chain and URDF serialized as contiguous text enclosed within boundary tokens (\texttt{<json\_start>},  \texttt{<json\_end>}, \texttt{<urdf\_start>}, \texttt{<urdf\_end>}).
At inference time, the model instead operates on point clouds segmented using our semi-automatic annotation strategy. Manual spatial prompts guide the extraction of a small number of articulation-relevant parts from raw scans or reconstructed meshes. The model then generates the Structural Reasoning Chain and URDF using the same prompting and decoding procedure as during training.

\section{Experiment}
\subsection{Experiments Settings}
\noindent \textbf{Experimental Setup.}
We train our model using the Adam optimizer with a fixed learning rate of $10^{-5}$ for 20 epochs. All experiments are conducted using 8$\times$ NVIDIA A100 GPUs. We adopt full-parameter finetuning of Qwen3-8B model. Mixed-precision training with \texttt{bfloat16} is enabled, and gradient checkpointing is used to reduce memory usage. Each batch consists of one single articulated object, where the full-object point cloud and per-part point clouds are encoded via a pretrained ULIP-2 (PointBERT) encoder with the Objaverse dataset and projected to the language model’s hidden space using a two-layer MLP. 
\begin{table*}[t]
\centering
\caption{Comparison of in-distribution(ID) joint estimation metrics.}
\label{tab:joint_metrics}
\resizebox{\linewidth}{!}{%
\begin{tabular}{lccccc}
\toprule
\textbf{Method} & \textbf{Type consistency (\%)} $\uparrow$ & \textbf{Axis position diff} $\downarrow$ & \textbf{Origin position diff} $\downarrow$ & \textbf{Angle limit diff} $\downarrow$ & \textbf{Distance limit diff} $\downarrow$ \\
\midrule
URDFormer &44.6 & 0.97 & 0.76 &0.82 &1.26 \\
Articulate-Anything & 59.0 & 0.27 & 0.32 &0.54 & 0.61 \\
\midrule
Arti4URDF & \textbf{94.9} & \textbf{0.05}    &  \textbf{0.15}   & \textbf{0.08}    & \textbf{0.14} \\
\bottomrule
\end{tabular}}
\vspace{-1em}
\end{table*}

\noindent \textbf{Dataset Construction.}
We leverage articulated objects from PartNet-Mobility and PhysXNet for training. PartNet-Mobility provides 2,346 instances over 46 categories, and we additionally select 8,400 articulated objects from the 26K-object PhysXNet dataset, resulting in roughly 11,000 objects in total. Each object is paired with a URDF file and fine-grained part-level semantic annotations. And we extract point clouds from each mesh model stored in PLY format with a global object and every part within the object.

\noindent\textbf{Evaluation Dataset.}
To better evaluate the performance and generalization ability of our model, we construct a comprehensive benchmark dataset that includes samples from PartNet-Mobility, PhysXNet, and a small set of real-world scanned objects. 
Specifically, we select 60 diverse instances from PartNet-Mobility and 60 from PhysXNet, ensuring coverage of all fundamental object categories for standard testing. 
These data can be seen as the in-distribution test data.
To assess out-of-distribution (OOD) generalization, we leave 100 objects from five out-of-distribution categories in the dataset: display, suitcase, plier, kettle and oven.
Then we apply 5 simulated room-scenes generated by OptiScene~\cite{yang2025optiscene} and 5 simulated table-scenes generated by MesaTask~\cite{hao2025mesatask}, total 10 scenes to test ArtiWold's capability.
Furthermore, we capture 10 real-world rigid objects with RGB images and reconstruct with Pi3~\cite{wang2025pi3}, to evaluate the model’s generalization to real, noisy, and incomplete inputs. This combined evaluation set provides a balanced and realistic benchmark for assessing both structural accuracy and real-world applicability.

\noindent \textbf{Evaluation Metrics.}
We evaluate the predicted URDF files with five metrics and provide full definitions and implementation details in the supplementary material:
\textbf{(1) Type Consistency (TC).}
Type consistency is the proportion of joints whose predicted type matches the ground truth type.
\noindent\textbf{(2) Axis Position Difference (APD).}
Axis position difference measures the average angular deviation between the predicted joint axis and the ground-truth joint axis.
\textbf{(3) Origin Position Difference (OPD).}
Origin position difference measures the average Euclidean distance between the predicted joint origin and the ground-truth joint origin.
\textbf{(4) Angle Limit Difference (ALD).}
Angle limit difference measures the average absolute error between the predicted and ground-truth rotational limits of a joint.
\textbf{(5) Distance Limit Difference (DLD).}
Distance limit difference measures the average absolute error between the predicted and ground-truth translational limits of a prismatic joint.
\noindent \textbf{(6) Articulated Object Usability Metrics.}
We assess the usability of predicted URDFs by matching predicted joints to ground-truth joints and then checking whether their parameters are sufficiently accurate. For each matched joint, we evaluate origin position, rotation axis alignment, and rotation range validity. An object is counted as usable only if all matched joints satisfy all three criteria, which provides a quantitative measure of whether the predicted URDF can be reliably used in physical simulation. 

\subsection{Compared Methods}
At the time of writing, the code and pre-trained models of \textit{URDF-Anything}~\cite{urdfanything} were not publicly available, so we do not include it as a baseline. Instead, we compare with two representative approaches: \textbf{URDFormer}~\cite{chen2024urdformer}, a vision-based pipeline that infers articulated structure from real-world RGB images and reconstructs simulator-ready objects with joint annotations, and \textbf{Articulated-Anything}~\cite{articulate-anything}, a language-guided framework that retrieves mesh segments and refines articulation proposals with an actor–critic loop to produce executable simulator code.

\subsection{Experimental Results}
\begin{figure*}
    \centering
    \includegraphics[width=0.94\linewidth]{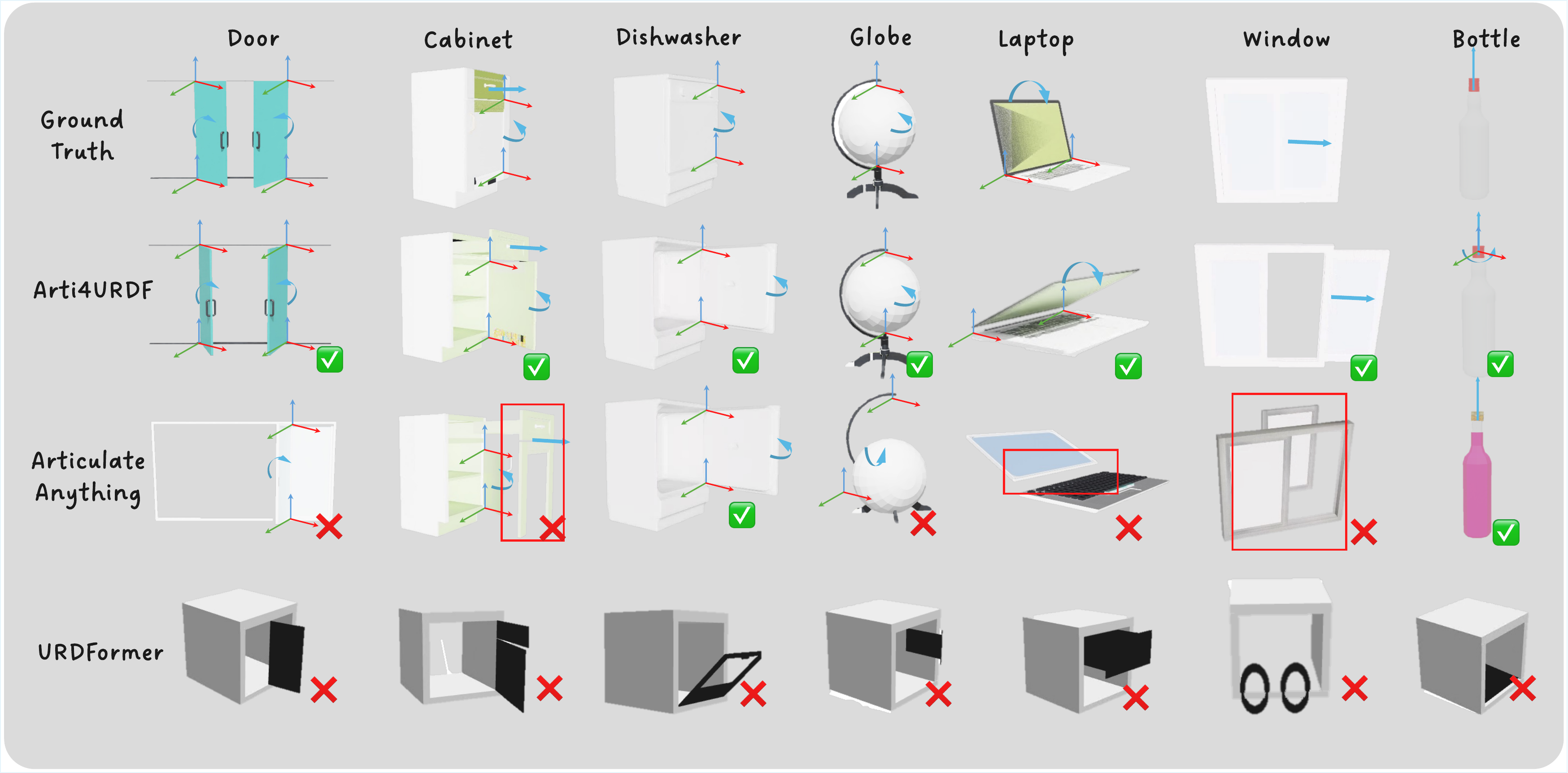}
    \vspace{-0.5em}
    \caption{
Qualitative comparison of articulated object reconstruction results.
}
    \label{fig:arti_results}
    \vspace{-1em}
\end{figure*}
\subsubsection{Reconstruction for in Sim-Objects}
\noindent \textbf{Qualitatively Results} We first present qualitative visualizations in~\cref{fig:arti_results} to highlight the quality of the articulated objects produced by our method. As shown in~\cref{fig:arti_results}, methods that rely on mesh or part retrieval, such as Articulate-Anything and URDFormer, often introduce noticeable changes to the original geometry. These approaches may replace object parts or generate simplified surrogate meshes, which leads to distortions in shape and appearance. In contrast, our method directly reasons over the original 3D geometry and predicts articulation based on the true structural relationships of the input asset. As a result, Arti4URDF preserves both the original shape and the visual characteristics of the object while producing accurate joint types and joint axes. The examples in the figure show that our outputs remain faithful to the input asset and recover articulation structure much closer to the ground truth.

\noindent \textbf{Joint Evaluation.} 
We evaluate the joint reconstruction quality under in-distribution (ID) settings using five key metrics: joint type consistency, axis position difference, origin position difference, angle limit difference, and distance limit difference. As shown in \cref{tab:joint_metrics}, our method Arti4URDF consistently outperforms both URDFormer and Articulate-Anything across all metrics by large margins.
Specifically, Arti4URDF achieves a type consistency of 94.9\%, significantly surpassing Articulate-Anything (59.0\%) and URDFormer (44.6\%). For the axis position difference, our method reduces the error to 0.05, a 5.4× improvement over Articulate-Anything (0.27) and 19× over URDFormer (0.97). Similarly, the origin position difference is lowered to 0.15, compared to 0.32 and 0.76 in the baselines. For joint movement constraints, our method attains the smallest angle limit difference (0.083) and distance limit difference (0.14), drastically outperforming URDFormer (0.82 / 1.26) and Articulate-Anything (0.54 / 0.61).

As shown in \cref{tab:ood}, our model also generalizes well to out-of-distribution (OOD) categories. Arti4URDF achieves a type consistency of \textbf{89.0\%}, which is more than twice that of URDFormer (23.8\%) and substantially higher than Articulate-Anything (43.5\%). For all joint-parameter metrics, our errors are also the lowest, with APD, OPD, ALD, and DLD reduced to \textbf{0.27}, \textbf{0.20}, \textbf{0.48}, and \textbf{0.34}, respectively. These results indicate that Arti4URDF maintains accurate joint prediction even on not seen classes.

\vspace{-0.1em}
\noindent \textbf{Articulated Object Usability Evaluation.}
To assess the practical utility of the predicted articulated objects, we evaluate not only joint-level accuracy but also whether the reconstructed object is usable as a whole. For this purpose, we compare the Arti4URDF with two baselines. As shown in \cref{tab:usability_results}, our method \textbf{Arti4URDF} achieves a significantly higher overall usability rate of \textbf{56.3\%}, compared to \textbf{35.2\%} for Articulate-Anything and only \textbf{18\%} for URDFormer. In both in-distribution (ID) and out-of-distribution (OOD) settings, Arti4URDF consistently outperforms prior baselines, showing strong generalization with \textbf{62.5\%} ID and \textbf{51.8\%} OOD success rates.

\begin{table}[t]
\centering
\caption{Results for out-of-distribution (OOD) evaluation}
\label{tab:ood}
\vspace{-0.5em}
\resizebox{\linewidth}{!}{%
\begin{tabular}{lccccc}
\toprule
\textbf{Method} &\textbf{TC} $\uparrow$ &\textbf{APD} $\downarrow$ & \textbf{OPD} $\downarrow$ & \textbf{ALD} $\downarrow$ & \textbf{DLD} $\downarrow$ \\
\midrule
URDFormer  & 23.8\% & 1.37 &0.92 & 1.63 & 0.87 \\
Articulate-Anything     & 43.5\% &0.48 &0.39 & 0.64& 0.66 \\
\midrule
Arti4URDF    & \textbf{89.0}\% & \textbf{0.27}&\textbf{0.20} & \textbf{0.48} & \textbf{0.34} \\
\bottomrule
\end{tabular}}
\vspace{-0.5em}
\end{table}

\begin{table}[t]
\centering
\small           
\setlength{\tabcolsep}{3pt}    
\renewcommand{\arraystretch}{0.9} 
\caption{Results for usability evaluation for objects.}
\label{tab:usability_results}
\vspace{-0.5em}
\begin{tabular}{lccc}
\toprule
\textbf{Method} & \textbf{ALL} & \textbf{ID}& \textbf{OOD}\\
\midrule
URDFormer  & 18.0\% & 24.6\% &8.3\% \\
Articulate-Anything    & 35.2\% &38.1\% &33.7\%  \\
\midrule
Arti4URDF      & \textbf{56.3\%} &\textbf{62.5\%} &\textbf{51.8\%} \\
\bottomrule
\end{tabular}
\vspace{-0.5em}
\end{table}

\begin{table}[t]
\centering
\small                         
\setlength{\tabcolsep}{3pt}    
\renewcommand{\arraystretch}{0.9} 
\caption{Results for sim scenes and real-world usability.}
\label{tab:sim_and_real_results}
\vspace{-0.5em}
\begin{tabular}{lcc|c}
\toprule
\textbf{Method} & \textbf{Sim-Id} & \textbf{Sim-Us} & \textbf{Real-world} \\
\midrule
URDFormer  & -& 6/25(24\%)&1/10(10\%)\\ 
Articulate-Anything    &-& 11/25(44\%)& 5/10(50\%) \\
\midrule
ArtiWorld     &\textbf{19/25(76\%)} & \textbf{15/25(60\%)}&\textbf{ 7/10(70\%)}\\
\bottomrule
\end{tabular}
\vspace{-2em}
\end{table}

\begin{figure*}[t]
    \centering
    \includegraphics[width=0.94\linewidth]{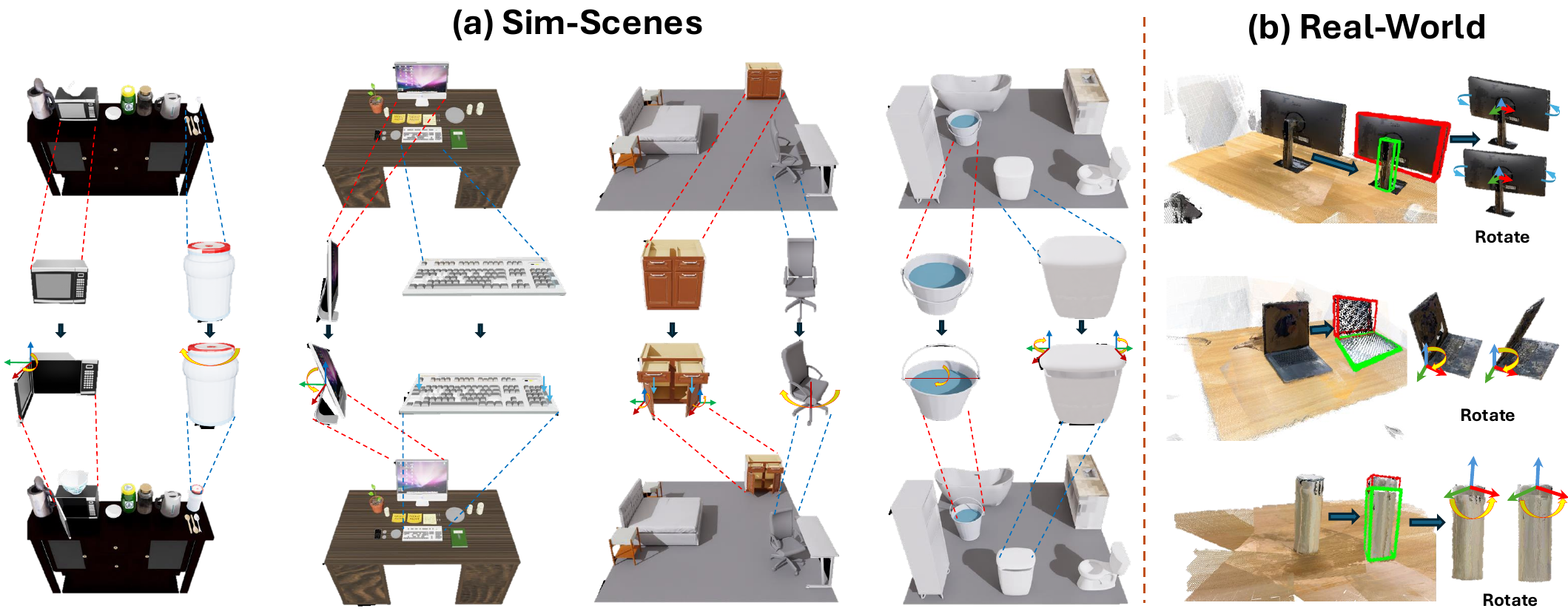}
    \vspace{-0.5em}
    \caption{
Qualitative results in simulated scenes and scanned real-world.
}
    \label{fig:simreal_results}
    \vspace{-1em}
\end{figure*}
\subsubsection{Reconstruction in Sim-Scenes}
As emphasized throughout this paper, the purpose of articulating objects is to enable better interaction between the robot and its environment in simulation. Building upon our object-level evaluations, we further assess whether an entire scene is usable as a whole. Specifically, for each scene, we evaluate (1) whether all articulable objects have been correctly identified (Sim-Id), and (2) whether all articulated objects in the scene are functional for using (Sim-Us). These criteria serve as the basis for assessing scene-level usability and we provide details in supplementary material. 
As shown in \cref{tab:sim_and_real_results}, the failures observed in Sim-Id are primarily attributed to ambiguities in textual matching when determining which objects within a scene should be classified as articulable. In cluttered or semantically diverse scenes, multiple objects may share similar descriptions, which introduces confusion in establishing ground-truth correspondences. Nonetheless, the high Sim-Us scores demonstrate that once an object is identified as articulable, our method is able to generate functional articulations that can be reliably used in simulation. This indicates that the core articulation-generation capability remains robust even when identification is imperfect.
\cref{fig:simreal_results}(a) provide qualitative visualizations of articulated objects in both tabletop and indoor-room scenes, further illustrating that our method produces coherent joint structures and enables realistic interaction within diverse environments.

\subsubsection{Reconstruction in Real-world}
Besides the above experiments, we test Arti4URDF on the common real-world scans 3D assets to show our model's generalization capability.
As shown in \cref{fig:simreal_results}(b), we provide a monitor, a laptop and a bottle as the real-world visualization examples with the well-articulated results. And from the right side of \cref{tab:sim_and_real_results}, only 1 out of 10 objects (10\%) from URDFormer are functional in real-world use, and Articulate-Anything improves this to 50\%. Our ArtiWorld achieves 70\% usability (7/10), significantly outperforming prior methods and demonstrating strong real-to-sim generalization.

\subsection{Ablation Studies}
In this section, we analyze the key components involved in the training pipeline and investigate their contributions.

\begin{table}[t]
\centering
\caption{Ablation Study on different factors of {Arti4URDF}.}
\label{tab:comprehensive_ablation}
\vspace{-0.5em}
\resizebox{\linewidth}{!}{
\begin{tabular}{lcccc}
\toprule
\textbf{Configuration} & \textbf{TC} $\uparrow$ & \textbf{OPD} $\downarrow$ & \textbf{ALD} $\downarrow$ & \textbf{DLD} $\downarrow$ \\
\midrule
\multicolumn{5}{l}{\textit{(a) Effect of LLM Scale}} \\
\quad Arti4URDF (Qwen3-1.7B) & 92.6 & 0.29 & 0.26 & 0.31 \\
\quad Arti4URDF (Qwen3-4B)   & 94.4 & 0.35 & 0.12 & 0.18 \\
\quad Arti4URDF (Qwen3-8B, Ours)   & \textbf{94.9} & \textbf{0.15} & \textbf{0.08} & \textbf{0.14} \\
\midrule
\multicolumn{5}{l}{\textit{(b) Effect of Training Data and 3D Structure Priors}} \\
\quad w/o PhysXNet, Structure P., Global PC & 89.2 & 0.39 & 0.24 & 0.28 \\
\quad + PhysXNet & 93.2 & 0.29 & 0.19 & 0.28 \\
\quad + Structure Prompts & 93.8 & 0.27 & 0.14 & 0.19 \\
\quad + Global PC (Ours) & \textbf{94.9} & \textbf{0.15} & \textbf{0.08} & \textbf{0.14} \\
\midrule
\multicolumn{5}{l}{\textit{(c) Effect of JSON Output Reasoning}} \\
\quad w/o JSON reasoning & 90.5 & 0.32 & 0.78 & 0.029 \\
\quad + JSON reasoning (Ours) & \textbf{94.9} & \textbf{0.15} & \textbf{0.08} & \textbf{0.14} \\
\bottomrule
\end{tabular}}
\vspace{-1.5em}
\end{table}

\noindent \textbf{Different versions of LLM.} To evaluate the reliability and generalization of our method, we test Arti4URDF on Qwen3 models of different scales. As shown in \cref{tab:comprehensive_ablation}(a), performance improves consistently as the model size grows. The 8B version achieves the highest type consistency and the lowest joint-parameter errors, indicating that larger LLMs offer stronger articulation reasoning.

\noindent \textbf{Only train with PartNet-Mobility.}
In \cref{tab:comprehensive_ablation}(b), we evaluate the effect of training data. When the model is trained only on PartNet-Mobility, its performance is significantly weaker under the same settings. Adding PhysXNet greatly improves type consistency and reduces all joint-parameter errors, showing that larger and more diverse articulated data are essential for learning reliable articulation behavior.

\noindent \textbf{3D Structure Priors.} Adding the global point cloud helps the model understand the overall geometry and the spatial relations among parts. We also include geometric relationship descriptions in the prompt to provide the LLM with a stronger spatial prior. As shown in \cref{tab:comprehensive_ablation}(b), removing both components leads to the weakest performance. Using only the structure prompt improves all metrics, and combining both achieves the best results, with higher type consistency and significantly lower joint-parameter errors.

\noindent\textbf{JSON reasoning.}
Our intermediate JSON output serves as a key CoT reasoning step that allows the model to first establish the geometric relationships between parts before predicting numerical joint parameters. In \cref{tab:comprehensive_ablation}(c), we evaluate the impact of removing this step. Without JSON reasoning, the model shows clear degradation in type consistency and errors in joint position and limits increase. With JSON reasoning, Arti4URDF achieves much higher accuracy across all metrics, demonstrating that structured intermediate reasoning is essential for precise articulation prediction.

\section{Conclusion}
We presented ArtiWorld, a framework that transforms non-articulated 3D assets into physically interactive articulated objects from scene descriptions. The core is Arti4URDF, an LLM-driven model that combines global and local 3D geometry to infer joint structures and generate full URDF specifications while preserving original object shape. Built on a large curated dataset from PartNet-Mobility and PhysXNet, ArtiWorld enables articulation at both object and scene level. Experiments show strong performance and generalization to unseen categories, outperforming existing alternatives. This work provides a scalable path toward articulated simulation and opens opportunities for richer robot interaction in complex environments.
{
    \small
    \bibliographystyle{ieeenat_fullname}
    \bibliography{main}
}


\end{document}